\pdfoutput=1

\def\year{2021}\relax
\documentclass[letterpaper]{article} 
\usepackage{aaai21}  
\usepackage{times}  
\usepackage{helvet} 
\usepackage{courier}  
\usepackage[hyphens]{url}  
\usepackage{graphicx} 
\usepackage{natbib}  
\usepackage{caption} 
\nocopyright
\usepackage{graphicx, amsmath, amssymb, amsthm, mathtools, nicefrac, algorithm, algorithmic, txfonts, centernot, xspace, subfig, adjustbox, dsfont}

\newtheorem{assum}{Assumption}[section]
\newtheorem{defn}{Definition}[section]

\newtheorem{thm}{Theorem}[section]
\newtheorem{lem}[thm]{Lemma}
\newtheorem{cor}[thm]{Corollary}

\newtheorem{rem}{Remark}[section]

\usepackage{tikz}
\usetikzlibrary{arrows, automata, positioning}

\newcommand{\Prob}{\mathbb{P}}
\newcommand{\Exp}{\mathbb{E}}
\newcommand{\Ind}{\mathds{1}}
\newcommand{\nPerp}{\centernot{\Perp}}

\DeclarePairedDelimiter{\floor}{\lfloor}{\rfloor}
\DeclarePairedDelimiter{\sbk}{[}{]}

\DeclarePairedDelimiter{\rbk}{(}{)}
\DeclarePairedDelimiter{\cbk}{\{}{\}}
\DeclarePairedDelimiter{\vbk}{|}{|}
\DeclarePairedDelimiter{\dvbk}{\|}{\|}

\newcommand\numberthis{\addtocounter{equation}{1}\tag{\theequation}}

\title{Loop Estimator for Discounted Values in Markov Reward Processes}
\author{
    Falcon Z.~Dai,
    Matthew R.~Walter \\
}
\affiliations{
    Toyota Technological Institute at Chicago\\
    Chicago, Illinois, USA 60637\\
    \{dai, mwalter\}@ttic.edu
}

\begin{document}


\maketitle

\begin{abstract}
At the working heart of policy iteration algorithms commonly used and studied in the discounted setting of reinforcement learning, the policy evaluation step estimates the value of states with samples from a Markov reward process induced by following a Markov policy in a Markov decision process.
We propose a simple and efficient estimator called \emph{loop estimator} that exploits the regenerative structure of Markov reward processes without explicitly estimating a full model. Our method enjoys a space complexity of $O(1)$ when estimating the value of a single positive recurrent state $s$ unlike TD with $O(S)$ or model-based methods with $O\left(S^2\right)$.
Moreover, the regenerative structure enables us to show, without relying on the generative model approach, that the estimator has an instance-dependent convergence rate of $\widetilde{O}\left(\sqrt{\tau_s/T}\right)$ over steps $T$ on a single sample path, where $\tau_s$ is the maximal expected hitting time to state $s$.
In preliminary numerical experiments, the loop estimator outperforms model-free methods, such as TD(k), and is competitive with the model-based estimator.

\end{abstract}

\section{Introduction}

The problem of policy evaluation arises naturally in the context of reinforcement learning (RL) \citep{sutton2018reinforcement} when one wants to evaluate the (action) values of a policy in a Markov decision process (MDP). In particular, policy iteration~\citep{howard1960dynamic} is a classic algorithmic framework for solving MDPs that poses and solves a policy evaluation problem during each iteration. Being motivated by the setting of reinforcement learning, i.e., the underlying MDP parameters are unknown and samples are obtained interactively, we focus on solving the policy evaluation problem given only a \emph{single} sample path.

Following a stationary Markov policy in an MDP, i.e., actions are determined based solely on the current state, gives rise to a \emph{Markov reward process} (MRP) \citep{puterman1994markov}. For the rest of the article, we focus on MRPs and consider the problem of estimating the infinite-horizon \emph{discounted} state values of an unknown MRP.

A straightforward approach to policy evaluation is to estimate the parameters of the MRP and then the value by plugging them into the classic Bellman equation \eqref{eq:immediate-bellman}~\citep{bertsekas1996neuro}. We call this the model-based estimator in the sequel. This approach is recently proved to be minimax-optimal given a generative model~\citep{Pananjady_Wainwright_2019} and it provides excellent estimates of discounted values in the single sample path setting as well, as our numerical experiments show (Section~\ref{sec:numerical}).
However, model-based estimators suffer from a space complexity of $O\rbk*{S^2}$, where $S$ is the number of states in the MRP. In contrast, \emph{model-free} methods enjoy a lower space complexity of $O(S)$ by not explicitly estimating the model parameters~\citep{sutton1988learning} but tend to exhibit a greater estimation error.

A popular class of estimators, $k$-step bootstrapping temporal difference or TD(k)%
\footnote{An important variant is TD($\lambda$), but we do not include it in our experiments since there is not a canonical implementation of the idea of estimating $\lambda$-return~\citep{sutton2018reinforcement}. However, any implementation is expected to exhibit similar behaviors as TD(k) with large $k$ corresponding to large $\lambda$~\citep{kearns2000bias}.} %
estimates a state's value based on the estimated values of other states. Like the model-based estimator, TD(k) is based on the classic Bellman equation~\eqref{eq:immediate-bellman}. The key property of the Bellman equation~\eqref{eq:immediate-bellman} is that the estimate of a state's value is tied to the estimates of other states which makes it hard to study the convergence of a specific state's value estimate in isolation and motivates the traditional approach of generative model in the literature.

Traditionally, prior works~\citep{kearns1999finite,even2003learning,Azar_Munos_Kappen_2013,Pananjady_Wainwright_2019} first show efficient estimation of \emph{all} state values under the assumption that we can generate a sample of next states and rewards starting in each states, and then invoke an argument that such a batch of samples can be obtained over a single sample path when all states are visited for at least once, i.e., over cover times.
In this work, we break with the traditional approach by directly studying the convergence of the value estimate of a \emph{single} state over the sample path. The convergence over all states is obtained as a simple consequence of the union bound.
Our key insight is that it is possible to circumvent the general difficulties of non-independent samples in the single sample path setting by recognizing the embedded regenerative structure of an MRP.
We alleviate the reliance on estimates of other states by studying segments of the sample path that start and end in the same state, i.e., \emph{loops}. This results in a novel and simple algorithm we call the \emph{loop estimator} (Algorithm~\ref{alg:loop}) which is a plug-in estimator based on a novel loop Bellman equation \eqref{eq:loop-bellman}. One important consequence is that the loop estimator can estimate the value of a single state with a space complexity of $O(1)$ which neither $TD(k)$ or the model-based estimator can achieve.

We first review the requisite definitions (Section~\ref{sec:preliminaries}) and then propose the loop estimator (Section~\ref{sec:loop-estimator}).
First, we analyze the algorithm's rate of convergence over visits to a single state (Theorem~\ref{thm:visits}).
Second, we study many steps it takes to visit a state. Using the exponential concentration of first return times (Lemma~\ref{lem:return-times}), we relate visits to their waiting times and establish the rate of convergence over steps (Theorem~\ref{thm:steps}).
Lastly, we obtain the convergence in $\ell_\infty$-norm over all states via the union bound as a consequence (Corollary~\ref{cor:inf-norm}).
Besides theoretical analysis, we also compare the loop estimator to several other estimators numerically on a commonly used example (Section~\ref{sec:numerical}).
Finally, we discuss the model-based vs. model-free status of the loop estimator (Section~\ref{sec:discussions}).

Our main contributions in this paper are two-fold:
\begin{itemize}
  \item By recognizing the embedded regenerative structure in MRPs, we derive a new Bellman equation over loops, segments that start and end in the same state.
  \item We introduce \emph{loop estimator}, a novel algorithm that can provably efficiently estimate the discounted values of a single state in an MRP from a single sample path.
\end{itemize}

In the interest of a concise presentation, we defer detailed proofs to Appendix~\ref{sec:proofs} with fully expanded logarithmic factors and constants. Similarly, see Appendix~\ref{sec:additional-results} for extra results. An implementation of the proposed loop estimator and presented experiments is publicly available.\footnote{\url{https://github.com/falcondai/loop-estimator}}

\section{Related works}

Much work that formally studies the convergence of value estimators (particularly the TD estimators) relies on having access to independent trajectories that start in \emph{all} states \citep{dayan1994td, even2003learning, jaakkola1994convergence, kearns2000bias}.
This is called a \emph{generative model} or sometimes a parallel sampling model \cite{kearns1999finite}. Given a convergence over batches of generative samples, we still need some reduction arguments to actually obtain a batch of generative (or parallel) samples over the sample path of a MRP.
\citet{kearns1999finite} consider how a set of independent trajectories can be obtained via mixing, i.e., approximately samples from the stationary distribution. This suggests on \emph{average} it takes $O\rbk*{t_\text{mix} / p^*}$-many steps where $t_\text{mix}$ is the expected steps to get close to the stationary distribution ($\nicefrac{1}{4}$ in total variation distance) and $p^*$ is the smallest probability in the stationary distribution.

This reduction can be improved by considering the steps the chain takes to visit all states at least once, i.e., \emph{cover times}, which is exactly when we have a batch of generative samples. This is an improved reduction in that we can study its convergence rate with high probability instead of the average behavior.
But the cover time of a Markov chain can be quite large: its concentration can be related to that of the hitting times to \emph{all} states. In contrast, for a single state, our results scale more favorably with the maximal expected hitting time of that state by a factor of $\log S$.
To ensure consistency of estimation is at all possible, we assume that the specific state to estimate is positive recurrent (Assumption~\ref{assum:reachability}), otherwise we cannot hope to (significantly) improve its value estimate after the final visit (see Appendix~\ref{sec:final-visit-example} for an illustrative example). We think that this assumption is reasonable as recurrence is a key feature of many Markov chains and it connects naturally to the online (interactive) setting where we cannot arbitrarily restart the chain. Moreover, this assumption is no stronger than the assumption used in the cover time reduction which assumes that we can repeatedly visit all states. If a resetting mechanism is available, values of transient states can be estimated from values of the recurrent states. Furthermore, in a finite MRP, there is at least one recurrent state due to the infinite length of a trajectory.

Besides the interest in the RL community to study the policy evaluation problem, operation researchers were also motivated to study estimation in order to leverage simulations as a computational tool. In such settings, the restriction of estimating only from a single sample path is usually not a concern.
Classic work in simulations by \citet{fox1989simulating} deals with estimating discounted value in a continuous time setting, including an estimator using regenerative structure. In comparison to their work, we provides an instance-dependent rate based on the transition structure which is relevant for the single sample path setting. \citet{Haviv_Puterman_1992} and \citet{derman1970finite} propose unbiased value estimators whereas the loop estimator is biased due to inversion.

Outside of the studies on reward processes, the regenerative structure of Markov chains has found application in the \emph{local} computation of PageRank~\citep{lee2013approximating}.
We make use of a lemma (Lemma~\ref{lem:return-times}, whose proof is included in the Appendix~\ref{sec:proof-lem-return-times} for completeness) from this work to establish an upper bound on waiting times (Corollary~\ref{cor:waiting-times}).
Furthermore, we provide an example to support why hitting times do not exponentially concentrate over its expectation in general (see Appendix~\ref{sec:hitting-time-example}).
Similar in spirit to the concept of locality studied by \citet{lee2013approximating}, our loop estimator enables space-efficient estimation of a single state value with a space complexity of $O(1)$ and an error bound without explicit dependency on the size of the state space. As a consequence, the loop estimator can provably estimate the value of a state with a finite maximal expected hitting time even if the state space is infinite.

Recently, an independent work by \citet{subramanian2019renewal} makes a similar observation of the regenerative structure and studies using estimates similar to the loop estimator in the context of a policy gradient algorithm. It provides promising experimental results that complement our novel theoretical guarantees on the rates of convergence. Taken together, these works show that regenerative structure is a promising direction in RL.

\section{Preliminaries}\label{sec:preliminaries}
\subsection{Markov reward processes and Markov chains}

Consider a finite state space $\mathcal{S} \coloneqq \{1, \cdots, S\}$ whose size is $S = \lvert \mathcal{S}\rvert$, %
a transition probability matrix $\mathbf{P} : \mathcal{S} \times \mathcal{S} \rightarrow [0, 1]$ that specifies the transition probabilities between consecutive states $X_t$ and $X_{t+1}$, %
i.e., (strong) Markov property $\Prob[X_{t+1}=s' \vert X_t = s, \cdots, X_0] = \Prob[X_{t+1}=s' \vert X_t=s] = P_{s s'} $, %
and a reward function $r : \mathcal{S} \rightarrow \mathcal{P}([0, r_\text{max}])$ where $R_t \sim r(X_t)$, then $(X_t, R_t)_{t \geq 0}$ is called a discrete-time finite \emph{Markov reward process} (MRP) \citep{puterman1994markov}.
Note that $(X_t)_{t \geq 0}$ is an embedded Markov chain with transition law $\mathbf{P}$. Furthermore, we denote the mean rewards as $\bar{\mathbf{r}} : s \mapsto \Exp[r(s)]$. As conventions, we denote $\Exp_s[\cdot] \coloneqq \Exp[\cdot \vert X_0 = s]$ and $\Prob_s[\cdot] \coloneqq \Prob[\cdot \vert X_0 = s]$.

The first step when a Markov chain visits a state $s$ is called the \emph{hitting time to $s$}, i.e., $H_s \coloneqq \inf \{ t : X_t = s \}$.
Note that if a chain starts at $s$, then $H_s = 0$. We refer to the first time a chain returns to $s$ as the \emph{first return time to $s$}
\begin{equation}\label{eq:first-return-time}
  H_s^+ \coloneqq \inf \{ t > 0 : X_t = s \}.
\end{equation}
\begin{defn}[Expected recurrence time]\label{def:expected-recurrence-time}
  Given a Markov chain, we define the \emph{expected recurrence time of state $s$} as the expected first return time of $s$ starting in $s$
  \begin{equation}\label{eq:expected-recurrence-time}
    \rho_s \coloneqq \Exp_s \left[ H_s^+ \right].
  \end{equation}
\end{defn}

A state $s$ is \emph{positive recurrent} if its expected recurrence time is finite, i.e., $\rho_s < \infty$.

\begin{defn}[Maximal expected hitting time]\label{def:maximal-expected-hitting-time}
  Given a Markov chain, we define the \emph{maximal expected hitting time of state $s$} as the maximal expected first return time over starting states
  \begin{equation}\label{eq:maximal-expected-hitting-time}
    \tau_s \coloneqq \max_{s' \in \mathcal{S}} \Exp_{s'} [ H_s^+ ].
  \end{equation}
\end{defn}

\subsection{Discounted total rewards}
In RL, we are generally interested in some expected long-term rewards that will be collected by following a policy. In the infinite-horizon discounted total reward setting, following a Markov policy on an MDP induces an MRP and the \emph{state value} of state $s$ is
\begin{equation}\label{eq:value}
  v(s) \coloneqq \Exp_s \left[ \sum_{t=0}^\infty \gamma^t R_t \right],
\end{equation}
where $\gamma \in [0, 1)$ is the \emph{discount factor}. Note that since the reward is bounded by $r_\text{max}$, state values are also bounded by $\nicefrac{r_\text{max}}{1 - \gamma}$.
A fundamental result relating values to the MRP parameters $(\mathbf{P}, \bar{\mathbf{r}})$ is the Bellman equation for each state $s \in \mathcal{S}$ \citep{sutton2018reinforcement}
\begin{equation}\label{eq:immediate-bellman}
  v(s) = \bar{r}_s + \gamma \sum_{s' \in \mathcal{S}} P_{s s'} v(s').
\end{equation}

%
\subsection{Problem statement}
Suppose that we have a sample path $(X_t, R_t)_{0 \leq t < T}$ of length $T$ from an MRP whose parameters $(\mathbf{P}, \bar{\mathbf{r}})$ are unknown. Given a state $s$ and discount factor $\gamma$, we want to estimate $v(s)$.

\begin{assum}[State $s$ is reachable]\label{assum:reachability}
  We assume state $s$ is reachable from all states, i.e., $\tau_s < \infty$.
\end{assum}

Otherwise there is some non-negligible probability that state $s$ will not be visited from some starting state. This will prevent the convergence in probability (in the form of a PAC-style error bound) that we seek (see Appendix~\ref{sec:final-visit-example}).
\begin{rem}\label{rem:reachability}
  Assumption~\ref{assum:reachability} can be weakened to the assumption that $s$ is positive recurrent and the MRP starts in the recurrent class containing $s$. All following results can be recovered by restricting $\mathcal{S}$ in the definition of $\tau_s$ to the recurrent class containing $s$. However, for ease of presentation, we will adopt Assumption~\ref{assum:reachability} in the rest of the article without loss of generality.
\end{rem}
Note that Assumption~\ref{assum:reachability} implies the positive recurrence of $s$, i.e., $\rho_s < \infty$, by definition, and that the MRP visits state $s$ for infinitely many times with probability 1.

%

\subsection{Renewal theory and loops}
Stochastic processes in general can exhibit complex dependencies between random variables at different steps, and thus often fall outside of the applicability of approaches that rely on independence assumptions. Renewal theory \citep{ross1996stochastic} focuses on a class of stochastic processes where the process restarts after a renewal event. Such regenerative structure allows us to apply results from the independent and identical distribution (IID) settings.

In particular, we consider the visits to state $s$ as renewal events and define \emph{waiting times} $W_n(s)$ for $n = 1, 2, \cdots$, to be the number of steps before the $n$-th visit
\begin{equation}\label{eq:waiting-times-def}
  W_n(s) \coloneqq \inf \left\{ w : n \leq \sum_{t=0}^w \Ind[ X_t = s ] \right\},
\end{equation}
and the \emph{interarrival times} $I_n(s)$ to be the steps between the $n$-th and $(n+1)$-th visit
\begin{equation}\label{eq:interarrival-times}
  I_n(s) \coloneqq W_{n+1}(s) - W_n(s).
\end{equation}

\begin{rem}\label{rem:times}
  The random times relate to each other in a few intuitive relations. The waiting time of the first visit is the same as the hitting time $W_1(s) = H_s \leq H_s^+$. Waiting times relate to interarrival times $W_{n+1}(s) = W_1(s) + \sum_{i=1}^n I_i(s)$.
\end{rem}
To justify treating visits to $s$ as renewal events, consider the sub-processes starting at $W_1(s)$ and at $W_2(s)$---both MRPs start in state $s$---due to Markov property of MRP, they are statistical replica of each other. Since segments $(X_t, R_t)_{W_n(s) \leq t < W_{n+1}(s)}$ start and end in the same state, we call them \emph{loops}. It follows that loops are independent of each other and obey the same statistical law. Intuitively speaking, an MRP is (probabilistically) specified by its starting state.

\begin{defn}[Loop $\gamma$-discounted rewards]
  Given a Markov reward process and a positive recurrent state $s$, we define the $n$-th \emph{loop $\gamma$-discounted rewards} as the discounted total rewards over the $n$-th loop
  \begin{equation}\label{eq:partial-sum}
    G_n(s) \coloneqq \sum_{u = 0}^{I_n(s) - 1} \gamma^u R_{W_n(s) + u}.
  \end{equation}
\end{defn}

\begin{defn}[Loop $\gamma$-discount]
  Given a Markov reward process and a positive recurrent state $s$, we define the $n$-th \emph{loop $\gamma$-discount} as the total discounting over the $n$-th loop
  \begin{equation}\label{eq:partial-discount}
    \Gamma_n(s) \coloneqq \gamma^{I_n(s)}.
  \end{equation}
\end{defn}

$\left(I_n(s), G_n(s)\right)_{n > 0}$ forms a regenerative process that has nice independence relations. Specifically, $I_n(s) \Perp I_m(s)$, $G_n(s) \Perp G_m(s)$, and $G_n(s) \Perp I_m(s)$ when $n \neq m$. Furthermore, $\rbk*{ I_n(s) }_{n > 0}$ are identically distributed the same as $H_s^+$ when starting in $s$. Similarly, $\rbk*{ G_n(s) }_{n > 0}$ are identically distributed. Note however that $G_n(s) \nPerp I_n(s)$.


\section{Main results}
\subsection{Bellman equations over loops}\label{sec:bellman}
Given the regenerative process $\left(I_n(s), G_n(s)\right)_{n > 0}$, we derive a new Bellman equation over the loops for state value $v(s)$.

\begin{thm}[Loop Bellman equations]\label{thm:loop-bellman}
  Suppose the expected loop $\gamma$-discount is $\alpha(s) \coloneqq \Exp_s[\Gamma_1(s)]$ and the expected loop $\gamma$-discounted rewards is $\beta(s) \coloneqq \Exp_s[G_1(s)]$, we can relate the state value $v(s)$ to itself
  \begin{equation}\label{eq:loop-bellman}
    v(s) = \beta(s) + \alpha(s) \, v(s).
  \end{equation}
\end{thm}

\begin{rem}
  The key difference between the loop Bellman equations \eqref{eq:loop-bellman} and the classic Bellman equations \eqref{eq:immediate-bellman} is the state values involved. Only state value $v(s)$ appears on the right-hand side of \eqref{eq:loop-bellman}.
\end{rem}

\subsection{Loop estimator}\label{sec:loop-estimator}
We plug in the empirical means for the expected loop $\gamma$-discount $\alpha(s)$ and the expected loop $\gamma$-discounted rewards $\beta(s)$ into the loop Bellman equation \eqref{eq:loop-bellman} and define the $n$-th \emph{loop estimator} for state value $v(s)$
\begin{equation}\label{eq:loop-estimator}
  \hat{v}_n(s) \coloneqq \hat{\beta}_n(s) / (1 - \hat{\alpha}_n(s)),
\end{equation}
where
%
%
  $\hat{\alpha}_n(s) \coloneqq \frac{1}{n} \sum_{i=1}^n \gamma^{I_i(s)}$
%
%
and
%
%
  $\hat{\beta}_n(s) \coloneqq \frac{1}{n} \sum_{i=1}^n G_i(s)$.
%
%
Furthermore, we have visited state $s$ for $(N+1)$ times before step $T$ where $N$ is a random variable that counts the number of loops before step $T$
\begin{equation}\label{eq:last-visit-time}
  N \coloneqq \sup \{ n : W_{n + 1}(s) \leq T \},
\end{equation}
and the estimate $\hat{v}_N(s)$ would be the last estimate before step $T$. Hence, with a slight abuse of notations, we define
\begin{equation}
  \hat{v}_T(s) \coloneqq \hat{v}_N(s).
\end{equation}
By using incremental updates to keep track of empirical means, Algorithm~\ref{alg:loop} implements the loop estimator $\hat{v}_T(s)$ with a space complexity of $O(1)$. Running $S$-many copies of loop estimators, one for each state $s \in \mathcal{S}$, takes a space complexity of $O(S)$.

\begin{algorithm}[!tb]
    \caption{Loop estimator (for a specific state)}
    \label{alg:loop}
    \begin{algorithmic}[1]
       \STATE {\bfseries Input:} discount factor $\gamma$, state $s$, sample path $(X_t, R_t)_{0 \leq t < T}$ of some length $T$.
       \STATE {\bfseries Return:} an estimate of the discounted value $v(s)$.

       \STATE Initialize the empirical mean of loop discounts $\hat{\alpha} \leftarrow 0$.
       \STATE Initialize the empirical mean of loop discounted rewards $\hat{\beta} \leftarrow 0$.
       \STATE Initialize the loop count $n \leftarrow 0$.
       \FOR {each loop in $(X_t, R_t)_{0 \leq t < T}$}
         \STATE Increment visit count $n \leftarrow n + 1$.
         \STATE Compute the length of the interarrival time $I_n(s) \leftarrow W_{n+1}(s) - W_n(s)$.
         \STATE Compute the partial discounted sum of rewards, $G_n(s) \leftarrow \sum_{u=0}^{I_n(s) - 1} \gamma^u R_{W_n(s) + u}$.
         \STATE Update the empirical means incrementally, $\hat{\alpha} \leftarrow \frac{1}{n} \gamma^{I_n(s)} + \left(1-\frac{1}{n} \right) \hat{\alpha}$, %
         and $\hat{\beta} \leftarrow \frac{1}{n} G_n(s) + \left(1-\frac{1}{n} \right) \hat{\beta}$.

       \ENDFOR
       \RETURN $\hat{\beta} / (1 - \hat{\alpha})$
    \end{algorithmic}
\end{algorithm}

\subsection{Rates of convergence}\label{sec:convergence-rates}
Now we investigate the convergence of the loop estimator, first over visits, i.e., $\hat{v}_n(s) \xrightarrow{p} v(s)$ as $n \rightarrow \infty$, then over steps, i.e., $\hat{v}_T(s) \xrightarrow{p} v(s)$ as $T \rightarrow \infty$.
By applying Hoeffding bound to the definition of loop estimator \eqref{eq:loop-estimator}, we obtain a PAC-style upper bound on the estimation error.

\begin{thm}[Convergence rate over visits]\label{thm:visits}
  Given a sample path from an MRP $(X_t, R_t)_{t \geq 0}$, a discount factor $\gamma \in [0, 1)$, and a positive recurrent state $s$, with probability of at least $1 - \delta$, the loop estimator converges to $v(s)$
  $$ | \hat{v}_n(s) - v(s) | = O\left( \frac{r_\text{max}}{ (1-\gamma)^2 } \sqrt{\frac{1}{n}\, \log \frac{1}{\delta} } \right). $$
\end{thm}

To determine the convergence rate over steps, we need to study the concentration of waiting times which allows us to lower-bound the random visits with high probability. As an intermediate step, we use the fact that the tail of the distribution of first return times is upper-bounded by an exponential distribution per the Markov property of MRP~\citep{lee2013approximating,aldous1999reversible}.

\begin{lem}[Exponential concentration of first return times \citep{lee2013approximating,aldous1999reversible}]\label{lem:return-times}
  Given a Markov chain $(X_t)_{t \geq 0}$ defined on a finite state space $\mathcal{S}$, for any state $s \in \mathcal{S}$ and any $t > 0$, we have
  $$ \Prob \left[ H_s^+ \geq t \right] \leq e \cdot e^{-t / e \tau_s}. $$
\end{lem}

Secondly, since by Remark~\ref{rem:times} we have $W_{n+1}(s) = W_1(s) + \sum_{i=1}^n I_i(s)$, we apply the union bound to upper-bound the tail of waiting times.

\begin{cor}[Upper bound on waiting times]\label{cor:waiting-times}
  With probability of at least $1 - \delta$, $W_n(s) = O\left(n\, \tau_s \, \log \frac{n}{\delta}\right)$.
\end{cor}

\begin{rem}
Note that the waiting time $W_n(s)$ is nearly linear in $n$ with a dependency on the Markov chain structure via the maximal expected hitting time of $s$, namely $\widetilde{O}(n\, \tau_s)$. In contrast, the \emph{expected} waiting time scales with the expected recurrence time $\Exp [W_n(s)] = \Theta(n\, \rho_s)$. However, an exponential concentration with the expected recurrence time is not possible in general (see Appendix~\ref{sec:hitting-time-example} for a counterexample).
\end{rem}

Using Lambert W function, we invert Corollary~\ref{cor:waiting-times} to lower-bound the visits by step $T$ with high probability. Finally, the convergence rate of $\hat{v}_T(s)$ follows from Theorem~\ref{thm:visits}.

\begin{thm}[Convergence rate over steps]\label{thm:steps}
  With probability of at least $1 - \delta$, for any $T > e \, \delta \, \tau_s$, the MRP $(X_t, R_t)_{0 \leq t < T}$ visits state $s$ for at least $\widetilde{\Omega}(T/\tau_s)$ many times, and the last loop estimate converges to $v(s)$
  $$ \vbk*{ \hat{v}_T(s) - v(s) } = \widetilde{O}\, \rbk*{ \frac{r_\text{max}}{(1-\gamma)^2} \sqrt{\frac{\tau_s}{T} \log \frac{1}{\delta}} } .$$
\end{thm}

Suppose we run a copy of loop estimator to estimate each state's value in $\mathcal{S}$, and denote them with a vector $\hat{\mathbf{v}}_T : s \mapsto \hat{v}_T(s)$. Convergence of the estimation error $\hat{\mathbf{v}}_T - \mathbf{v}$ in terms of the $\ell_\infty$-norm follows immediately by applying the union bound.

\begin{cor}[Convergence rate over all states]\label{cor:inf-norm}
  With probability of at least $1 - \delta$, for any $T > e \, \delta \, \max_s \tau_s$, the MRP $(X_t, R_t)_{0 \leq t < T}$ visits each state $s$ for at least $\widetilde{\Omega}(T/\tau_s)$ many times, and the last loop estimates converge to state values $\mathbf{v}$
  $$ \dvbk{ \hat{\mathbf{v}}_T - \mathbf{v} }_\infty = \widetilde{O}\, \rbk*{ \frac{r_\text{max}}{(1-\gamma)^2} \sqrt{\frac{\max_s \tau_s}{T} \log \frac{S}{\delta}} } .$$
\end{cor}

\section{Numerical experiments}\label{sec:numerical}
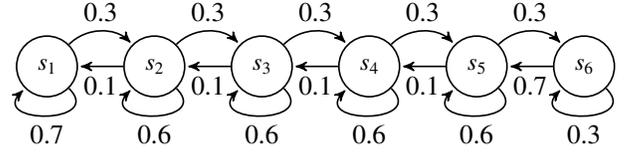
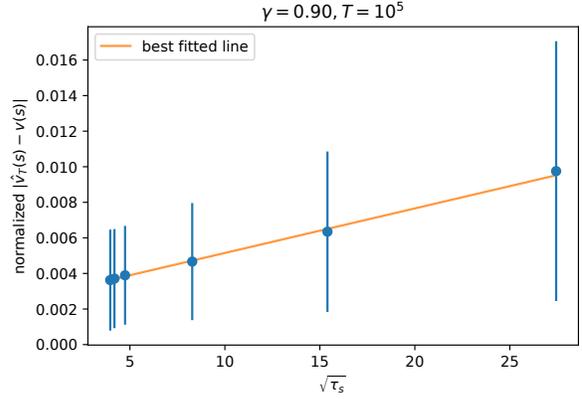
\begin{figure}[!tb]
  \centering
  \subfloat[]{%
    \label{fig:river-swim}%
    \raisebox{-0.5\height}{%
      \resizebox{0.49\textwidth}{!}{%
        \begin{tikzpicture}[->, >=stealth', shorten >=1pt, auto, semithick]
          \tikzstyle{action} = [draw=black,fill=none]

            \node[state, scale=0.9] at (-7/5*3, 0) (s1) {$s_1$};
            \node[state, scale=0.9] at (-7/5*2, 0) (s2) {$s_2$};
            \node[state, scale=0.9] at (-7/5*1, 0) (s3) {$s_3$};
            \node[state, scale=0.9] at (7/5*0, 0) (s4) {$s_4$};
            \node[state, scale=0.9] at (7/5*1, 0) (s5) {$s_5$};
            \node[state, scale=0.9] at (7/5*2, 0) (s6) {$s_6$};

            %

            \path (s1) edge [out=45, in=135] node {$0.3$} (s2)
                       edge [out=-45,in=-135,looseness=3] node {$0.7$} (s1);
            \path (s2) edge [out=45, in=135] node {$0.3$} (s3)
                       edge [out=-45,in=-135,looseness=3] node {$0.6$} (s2)
                       edge [out=-180, in=0] node {$0.1$} (s1);
            \path (s3) edge [out=45, in=135] node {$0.3$} (s4)
                       edge [out=-45,in=-135,looseness=3] node {$0.6$} (s3)
                       edge [out=-180, in=0] node {$0.1$} (s2);
            \path (s4) edge [out=45, in=135] node {$0.3$} (s5)
                       edge [out=-45,in=-135,looseness=3] node {$0.6$} (s4)
                       edge [out=-180, in=0] node {$0.1$} (s3);
            \path (s5) edge [out=45, in=135] node {$0.3$} (s6)
                       edge [out=-45,in=-135,looseness=3] node {$0.6$} (s5)
                       edge [out=-180, in=0] node {$0.1$} (s4);
            \path (s6) edge [out=-45,in=-135,looseness=3] node {$0.3$} (s6)
                       edge [out=-180, in=0] node {$0.7$} (s5);
        \end{tikzpicture}%
      }%
    }%
  }%
  \hfill
  \subfloat[]{%
    \raisebox{-0.5\height}{%
      \label{fig:loop-states}%
      \includegraphics[width=\columnwidth]{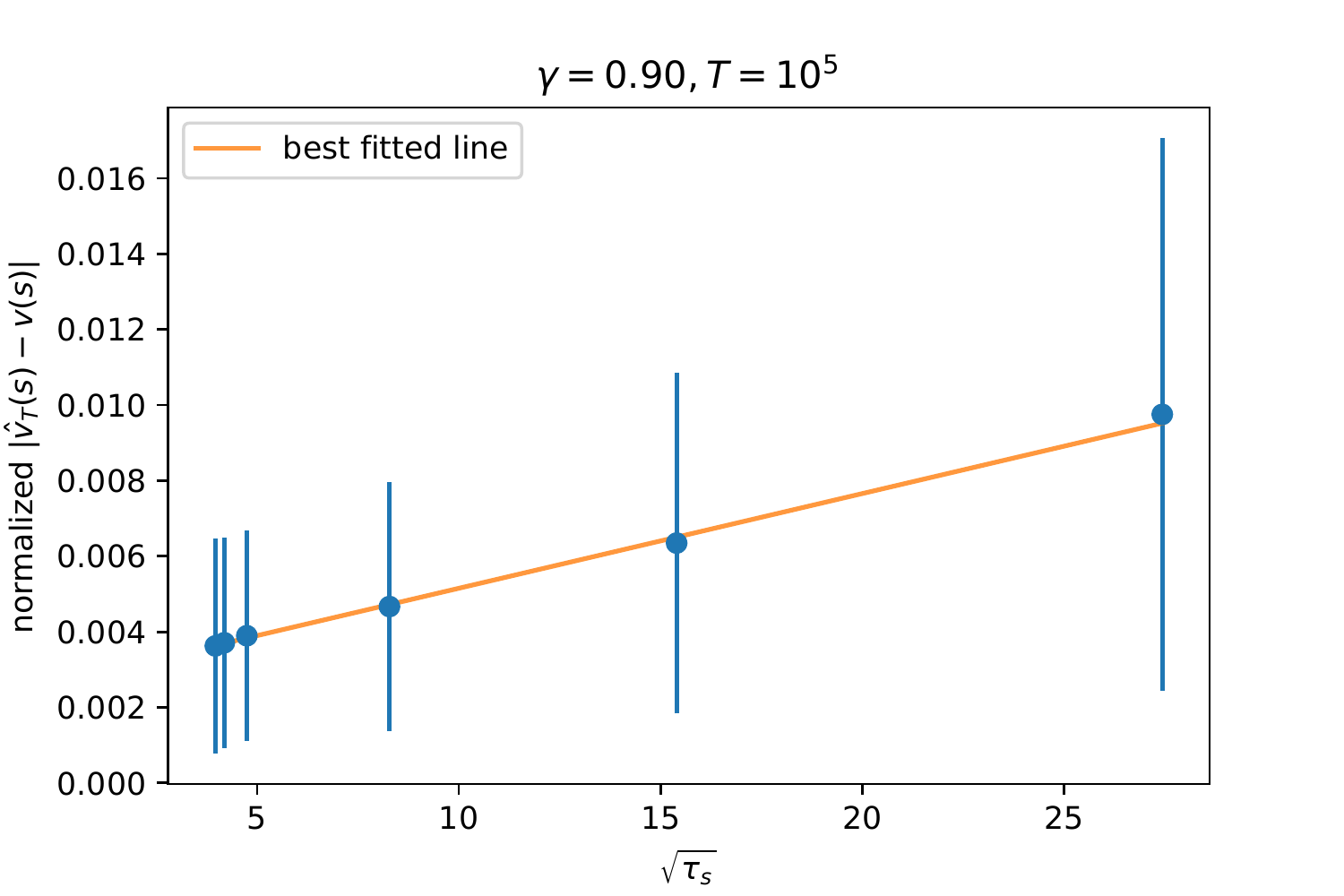}%
    }%
  }%
  \caption[]{\subref{fig:river-swim} The induced \texttt{RiverSwim} MRP. The arrows are labeled with transition probabilities. The rewards are all zero except for state $s_6$, where $r(s_6) = 1$.
  \subref{fig:loop-states} With discount factor $\gamma=0.9, T=10^5$. The estimation error of each state (normalized by $\max_s v(s)$) is plotted over the square root of maximal expected hitting times $\sqrt{\tau_s}$ of that state. Error bars show the standard deviations over 200 runs.}%
\end{figure}

We consider \texttt{RiverSwim}, an MDP proposed by \citet{strehl2008analysis} that is often used to illustrate the challenge of exploration in RL. The MDP consists of six states $\mathcal{S} = \{ s_1, \cdots, s_6 \}$ and two actions $\mathcal{A} = \{\text{``swim downstream''}, \text{``swim upstream''} \}$. Executing the ``swim upstream'' action often fails due to the strong current, while there is a high reward for staying in the most upstream state $s_6$. For our experiments, we use the MRP induced by always taking the ``swim upstream'' action (see Figure~\ref{fig:river-swim} for numerical details).

The most relevant aspect of the induced MRP is that the maximal expected hitting times are very different for different states: %
$\tau_{s_1} \approx 752$, $\tau_{s_2} \approx 237$, $\tau_{s_3} \approx 68$, $\tau_{s_4} \approx 15$, $\tau_{s_5} \approx 17$, $\tau_{s_6} \approx 22$.
Figure~\ref{fig:loop-states} shows a plot of the estimation errors of the loop estimator for each state over the \emph{square root} of maximal expected hitting times $\sqrt{\tau_s}$ of that state. The observed linear relationship between the two quantities (supported by a good linear fit) is consistent with the instance-dependence in our result of $\vbk*{ \hat{v}_T(s) - v(s) } = \widetilde{O}\rbk*{\sqrt{\tau_s}}$, c.f., Theorem~\ref{thm:steps}.

\subsection{Alternative estimators}
We define several alternative estimators for $v(s)$ and briefly mention their relevance for comparison.

{\bf Model-based.} We compute add-1 smoothed maximum likelihood estimates (MLE) of the MRP parameters $\left(\mathbf{P}, \bar{\mathbf{r}}\right)$ from the sample path
\begin{equation}\label{eq:model-based-p}
  \hat{P}_{s\,s'} \coloneqq \frac{\frac{1}{S} + \sum_{t=0}^{T-1} \Ind \left[ X_{t+1} = s', X_t = s \right]}{1 + \sum_{t=0}^{T-1} \Ind \left[ X_t = s \right]}
\end{equation}
and
\begin{equation}\label{eq:model-based-r}
  \hat{\bar{r}}_s \coloneqq \frac{\sum_{t=0}^{T-1} R_t \Ind \left[ X_t = s \right]}{1 + \sum_{t=0}^{T-1} \Ind \left[ X_t = s \right]}.
\end{equation}
We then solve for the discounted state values from the Bellman equation \eqref{eq:immediate-bellman} for the MRP parameterized by $\left( \hat{\mathbf{P}}, \hat{\bar{\mathbf{r}}} \right)$, i.e., the (column) vector of estimated state values
\begin{equation}\label{eq:model-based}
  \hat{\mathbf{v}}_\text{MB} \coloneqq \left( \mathbf{I} - \gamma \hat{\mathbf{P}} \right)^{-1} \hat{\bar{\mathbf{r}}}
\end{equation}
where $\mathbf{I}$ is the identity matrix.

{\bf TD(k).} $k$-step temporal difference (or $k$-step backup) estimators are commonly recursively defined \citep{kearns2000bias} with TD(0) being a textbook classic \citep{bertsekas1996neuro,sutton2018reinforcement}. Let
$\hat{v}_\text{TD}(0, s) \coloneqq 0$
for all states $s \in \mathcal{S}$. And for $t > 0$
\begin{equation*}\label{eq:td-k}
  \hat{v}_\text{TD}(t, s) \coloneqq \left\{
    \begin{aligned}
      & (1 - \eta_t)\, \hat{v}_\text{TD}(t-1, s) \\
      &\quad + \eta_t \Big( \gamma^0 R_t + \cdots + \gamma^{k} R_{t+k} \\
      &\quad + \gamma^{k+1} \hat{v}_\text{TD}(t-1, X_{t+k+1}) \Big), & \text{if }s = X_t & \\
      & \hat{v}_\text{TD}(t-1, s), & \text{otherwise} &
    \end{aligned}
  \right.
\end{equation*}
where $\eta_t$ is the learning rates. A common choice is to set $\eta_t = 1 / \rbk*{ \sum_{u=0}^{t} \Ind \sbk*{ X_u = s } }$ which satisfies the Robbins-Monro conditions \citep{bertsekas1996neuro}.
But it has been shown to lead to slower convergence than $\eta_t = 1 / \rbk*{ \sum_{u=0}^{t} \Ind \sbk*{ X_u = s } }^d$ where $d \in (\nicefrac{1}{2}, 1)$ \citep{even2003learning}.

It is more accurate to consider TD methods as a large family of estimators each with different choices of $k$, $\eta_t$. Choosing these parameters can create extra work and sometimes confusion for practitioners. Whereas the loop estimator, like the model-based estimator, has no parameters to tune. In any case, it is not our intention to compare with the TD family exhaustively (see more results on TD on \citep{kearns2000bias,even2003learning}). Instead, we will compare with $\text{TD}(0)$ and $\text{TD}(10)$, both with $d = 1$, and $\text{TD}(0)^*$ with $d = \nicefrac{1}{2}$.

\subsection{Comparative experiments}
\begin{figure}[!htb]
  \centering
    \includegraphics[width=\columnwidth]{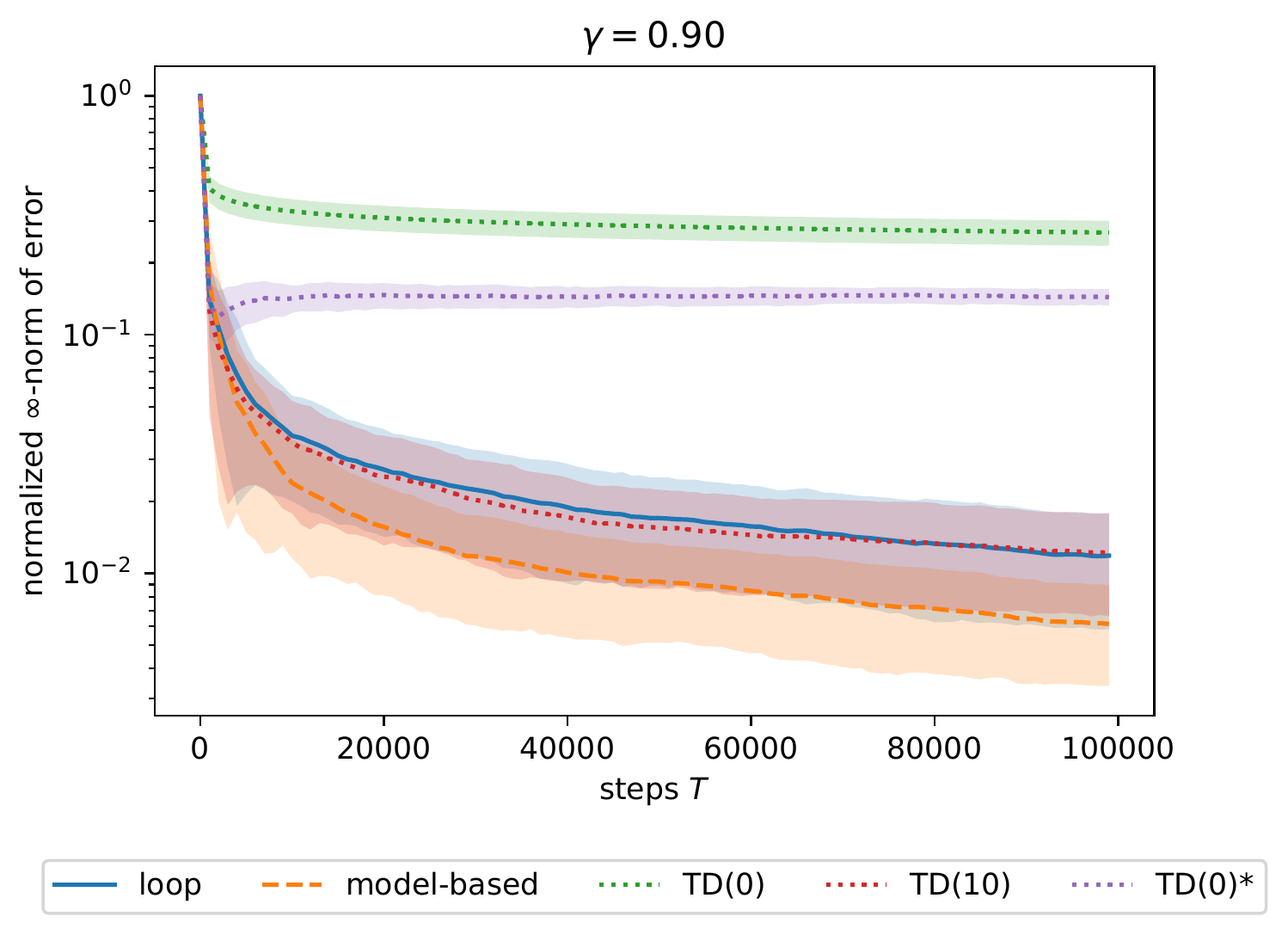}
  \hfill
    \includegraphics[width=\columnwidth]{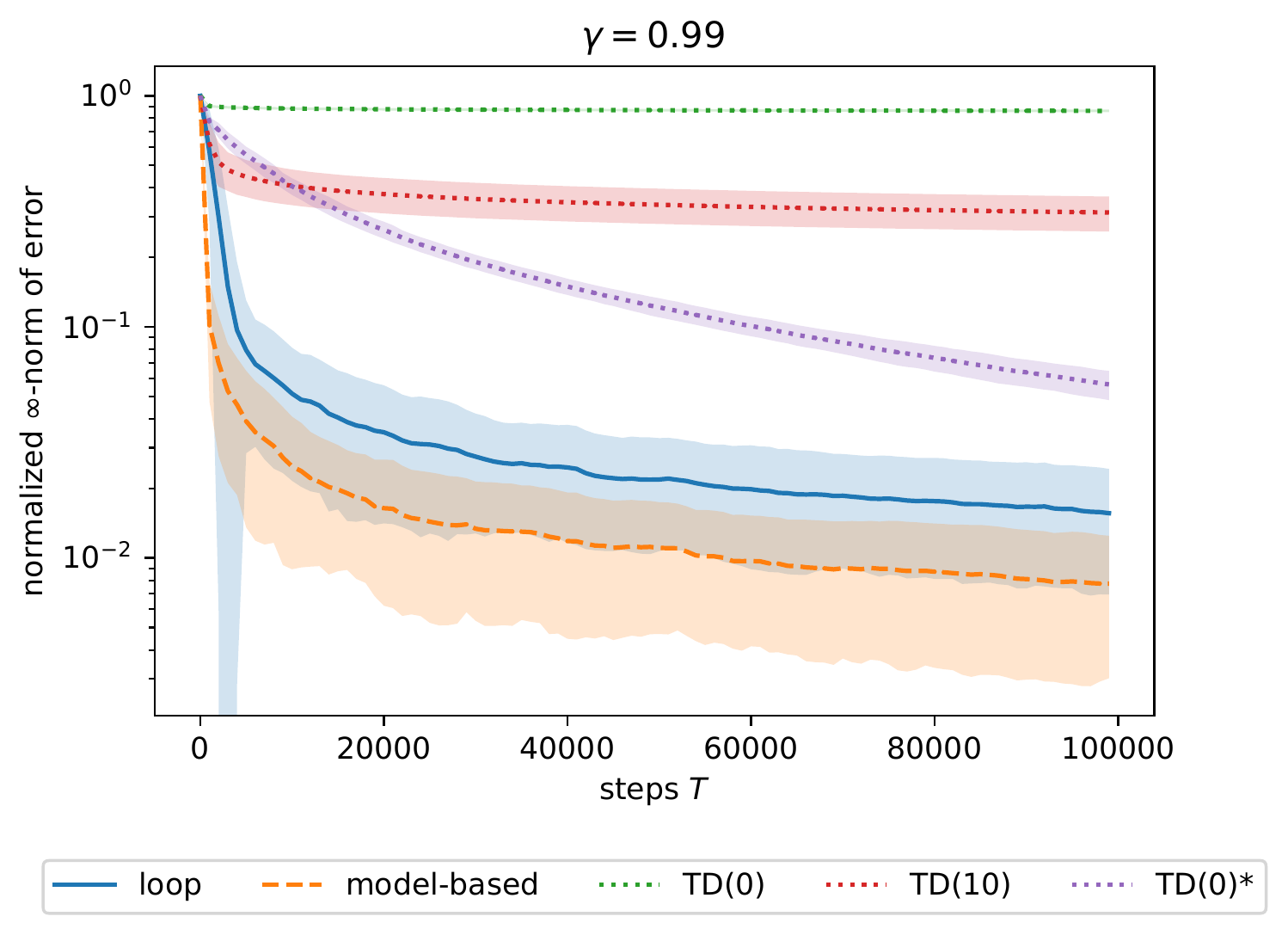}
  \caption{Estimation errors (normalized by $\max_s v(s)$ to be comparable across discount factors) of different estimators at different discount factors (left) $\gamma=0.9$ and (right) $\gamma=0.99$. Shaded areas represent the standard deviations over 200 runs. Note the vertical log scale.}
  \label{fig:gammas}
\end{figure}

We experiment with different values for the discount factor $\gamma$, because, roughly speaking, $1 / (1-\gamma)$ sets the horizon beyond which rewards are discounted too heavily to matter. We compare the estimation errors measured in $\infty$-norm, which is important in RL. The results are shown in Figure~\ref{fig:gammas}.
\begin{itemize}
  \item The model-based estimator dominates all estimators for every discount setting we tested.
  \item TD(k) estimators perform well if $k \geq 1/(1-\gamma)$.
  \item The loop estimator performs worse than, but is competitive with, the model-based estimator. Furthermore, similar to the model-based estimator and unlike the TD(k) estimators, its performance seems to be less influenced by discounting.
\end{itemize}


\section{Discussions}\label{sec:discussions}
The elementary identity below relates the expected first return times $Y_{s\,s'} \coloneqq \Exp_s \left[ H_{s'}^+ \right]$ to the transition probabilities $P_{s\, s'}$ for a finite Markov chain. Using the matrix notations, suppose that the expected first return times are organized in a matrix $\mathbf{Y}$, and $\mathbf{P}$ the transition matrix of the Markov chain, then we have
$ \mathbf{Y} = \mathbf{P}\left( \mathbf{Y} - \text{diag} \mathbf{Y} + \mathbf{E} \right) $
where $\text{diag} \mathbf{Y}$ is a matrix with the same diagonal as $\mathbf{Y}$ and zero elsewhere, and $\mathbf{E}$ is a matrix with all ones.
Thus, knowing $\mathbf{Y}$ is equivalent to knowing the full model, as we can compute $\mathbf{P}$ using this identity.
Recall that by definition $\Exp\left[ I_1(s) \right] = \Exp_s \left[ H_s^+ \right]$, which is exactly the diagonal of $\mathbf{Y}$.
But only knowing the diagonal is not sufficient to determine the entire set of model parameters, namely $\mathbf{Y}$, the loop estimator based on $\rbk*{I_n}_{n > 0}$ indeed falls short of being a model-based method. It may be considered a \emph{semi}-model-based method as it estimates some but not all of the model parameters.


For large MRPs, a natural extension of our work is to consider recurrence of features instead of states, e.g., a video game screen might not repeat itself completely but the same items might reappear. After all, without repetition exactly or approximately, it would not be possible for an agent to learn and improve its decisions.

We believe that regenerative structure can be further exploited in RL (particularly in the form of the loop Bellman equation \eqref{eq:loop-bellman}) and we think this article provides the fundamental results for future study in this direction.

\section*{Acknowledgments}

This work was supported in part by the National Science Foundation under Grant No.~1830660. We thank Mesrob I.~Ohannessian for a helpful discussion on Markov chains, and Christina Lee Yu for discussing an early version of this work.
We also thank anonymous reviewers for their constructive feedback, in particular, for bringing an independent work \cite{subramanian2019renewal} to our attention.

\bibliography{references}

\begin{thebibliography}{20}
\providecommand{\natexlab}[1]{#1}
\providecommand{\url}[1]{\texttt{#1}}
\providecommand{\urlprefix}{URL }
\expandafter\ifx\csname urlstyle\endcsname\relax
  \providecommand{\doi}[1]{doi:\discretionary{}{}{}#1}\else
  \providecommand{\doi}{doi:\discretionary{}{}{}\begingroup
  \urlstyle{rm}\Url}\fi

\bibitem[{Aldous and Fill(1999)}]{aldous1999reversible}
Aldous, D.; and Fill, J. 1999.
\newblock Reversible {M}arkov chains and random walks on graphs.
\newblock Book in preparation (available at
  \url{http://www.stat.berkeley.edu/~aldous/RWG/Chap2.pdf}).

\bibitem[{Bertsekas and Tsitsiklis(1996)}]{bertsekas1996neuro}
Bertsekas, D.~P.; and Tsitsiklis, J.~N. 1996.
\newblock \emph{Neuro-dynamic programming}.
\newblock Athena Scientific Belmont, MA.

\bibitem[{Dayan and Sejnowski(1994)}]{dayan1994td}
Dayan, P.; and Sejnowski, T.~J. 1994.
\newblock {TD} ($\lambda$) converges with probability 1.
\newblock \emph{Machine Learning} 14(3): 295--301.

\bibitem[{Derman(1970)}]{derman1970finite}
Derman, C. 1970.
\newblock \emph{Finite state {M}arkovian decision processes}.
\newblock Academic Press.

\bibitem[{Even-Dar and Mansour(2003)}]{even2003learning}
Even-Dar, E.; and Mansour, Y. 2003.
\newblock Learning rates for {Q}-learning.
\newblock \emph{Journal of machine learning Research} 5(Dec): 1--25.

\bibitem[{Fox and Glynn(1989)}]{fox1989simulating}
Fox, B.~L.; and Glynn, P.~W. 1989.
\newblock Simulating discounted costs.
\newblock \emph{Management Science} 35(11): 1297--1315.

\bibitem[{Gheshlaghi~Azar, Munos, and Kappen(2013)}]{Azar_Munos_Kappen_2013}
Gheshlaghi~Azar, M.; Munos, R.; and Kappen, H.~J. 2013.
\newblock Minimax PAC bounds on the sample complexity of reinforcement learning
  with a generative model.
\newblock \emph{Machine Learning} 91(3): 325–349.
\newblock ISSN 1573-0565.
\newblock \doi{10.1007/s10994-013-5368-1}.

\bibitem[{Haviv and Puterman(1992)}]{Haviv_Puterman_1992}
Haviv, M.; and Puterman, M.~L. 1992.
\newblock Estimating the value of a discounted reward process.
\newblock \emph{Operations Research Letters} 11(5): 267–272.
\newblock ISSN 01676377.
\newblock \doi{10.1016/0167-6377(92)90002-K}.

\bibitem[{Howard(1960)}]{howard1960dynamic}
Howard, R.~A. 1960.
\newblock \emph{Dynamic programming and {M}arkov processes.}
\newblock John Wiley.

\bibitem[{Jaakkola, Jordan, and Singh(1994)}]{jaakkola1994convergence}
Jaakkola, T.; Jordan, M.~I.; and Singh, S.~P. 1994.
\newblock Convergence of stochastic iterative dynamic programming algorithms.
\newblock In \emph{Advances in Neural Information Processing Systems},
  703--710.

\bibitem[{Kearns and Singh(1999)}]{kearns1999finite}
Kearns, M.~J.; and Singh, S.~P. 1999.
\newblock Finite-sample convergence rates for {Q}-learning and indirect
  algorithms.
\newblock In \emph{Advances in Neural Information Processing Systems},
  996--1002.

\bibitem[{Kearns and Singh(2000)}]{kearns2000bias}
Kearns, M.~J.; and Singh, S.~P. 2000.
\newblock Bias-Variance Error Bounds for Temporal Difference Updates.
\newblock In \emph{Conference on Learning Theory}, 142--147.

\bibitem[{Lee, Ozdaglar, and Shah(2013)}]{lee2013approximating}
Lee, C.~E.; Ozdaglar, A.; and Shah, D. 2013.
\newblock Approximating the Stationary Probability of a Single State in a
  {M}arkov chain.
\newblock \emph{arXiv preprint}
  \urlprefix\url{https://arxiv.org/abs/1312.1986}.

\bibitem[{Pananjady and Wainwright(2019)}]{Pananjady_Wainwright_2019}
Pananjady, A.; and Wainwright, M.~J. 2019.
\newblock Value function estimation in Markov reward processes:
  Instance-dependent $\ell_\infty$-bounds for policy evaluation.
\newblock \emph{arXiv preprint}
  \urlprefix\url{https://arxiv.org/abs/1909.08749}.

\bibitem[{Puterman(1994)}]{puterman1994markov}
Puterman, M.~L. 1994.
\newblock \emph{Markov Decision Processes: {D}iscrete Stochastic Dynamic
  Programming}.
\newblock John Wiley \& Sons, Inc.

\bibitem[{Ross(1996)}]{ross1996stochastic}
Ross, S.~M. 1996.
\newblock \emph{Stochastic processes}.
\newblock John Wiley, 2nd edition.

\bibitem[{Strehl and Littman(2008)}]{strehl2008analysis}
Strehl, A.~L.; and Littman, M.~L. 2008.
\newblock An analysis of model-based interval estimation for {M}arkov decision
  processes.
\newblock \emph{Journal of Computer and System Sciences} 74(8): 1309--1331.

\bibitem[{{Subramanian} and {Mahajan}(2019)}]{subramanian2019renewal}
{Subramanian}, J.; and {Mahajan}, A. 2019.
\newblock Renewal Monte Carlo: Renewal theory based reinforcement learning.
\newblock \emph{IEEE Transactions on Automatic Control} 1--1.

\bibitem[{Sutton(1988)}]{sutton1988learning}
Sutton, R.~S. 1988.
\newblock Learning to predict by the methods of temporal differences.
\newblock \emph{Machine Learning} 3(1): 9--44.

\bibitem[{Sutton and Barto(2018)}]{sutton2018reinforcement}
Sutton, R.~S.; and Barto, A.~G. 2018.
\newblock \emph{Reinforcement learning: {A}n introduction}.
\newblock MIT press, 2nd edition.

\end{thebibliography}

\clearpage
\appendix
\allowdisplaybreaks
\section{Detailed proofs}
\label{sec:proofs}

\subsection{Proof of Theorem~\ref{thm:loop-bellman}}
\label{sec:proof-loop-bellman}

\begin{proof}
  Note that since $X_0 = s$, we have $W_1(s) = 0$ and $I_1(s) = W_2(s)$. Since only state $s$ appears here, we will suppress $s$ from the random variables below to simplify the notation. We use Assumption~\ref{assum:reachability} or the weaker assumption that $s$ is positive recurrent, i.e., $\rho_s < \infty$, to guarantee that $W_2 < \infty$ with probability 1.

  \begin{align*}
    &v(s) \\
    &\quad\triangleright\text{By definition \eqref{eq:value}} \\
    &\quad= \Exp_s \left[ \sum_{t=0}^\infty \gamma^t R_t \right] \\
    &\quad\triangleright\text{Split the infinite sum at $W_2$} \\
    &\quad= \Exp_s \left[ \sum_{t=0}^{W_2 - 1} \gamma^t R_t + \sum_{t=W_2}^\infty \gamma^t R_t \right] \\
    &\quad\triangleright\text{Rewrite the indices} \\
    &\quad= \Exp_s \left[ \sum_{t=0}^{W_2 - 1} \gamma^t R_{W_1 + t} + \gamma^{W_2} \sum_{t=0}^\infty \gamma^t R_{W_2 + t} \right] \\
    &\quad\triangleright\text{By definition \eqref{eq:interarrival-times}} \\
    &\quad= \Exp_s \left[ \sum_{t=0}^{I_1 - 1} \gamma^t R_{W_1 + t} + \gamma^{I_1} \sum_{t=0}^\infty \gamma^t R_{W_2 + t} \right] \\
    &\quad\triangleright\text{By definition \eqref{eq:partial-sum}} \\
    &\quad= \Exp_s \left[ G_1 + \gamma^{I_1} \sum_{t=0}^\infty \gamma^t R_{W_2 + t} \right] \\
    &\quad\triangleright\text{Split off the first term} \\
    &\quad= \Exp_s \left[ G_1 \right] + \Exp_s \left[ \gamma^{I_1} \sum_{t=0}^\infty \gamma^t R_{W_2 + t} \right] \\
    &\quad\triangleright\text{Note that $X_{W_2} = s$ and by Markov property} \\
    &\quad= \Exp_s \left[ G_1 \right] + \Exp_s \left[ \gamma^{I_1} \right] \Exp_s \left[ \sum_{t=0}^\infty \gamma^t R_{W_2 + t} \right] \\
    &\quad\triangleright\text{$(R_{W_2+t})_{t \geq 0}$ and $(R_t)_{t \geq 0}$ are probabilistically identical} \\
    &\quad= \Exp_s \left[ G_1 \right] + \Exp_s \left[ \gamma^{I_1} \right] v(s) \\
    &\quad\triangleright\text{By the definitions of $\alpha(s)$ and $\beta(s)$}\\
    &\quad= \beta(s) + \alpha(s) \, v(s). &&\qedhere
  \end{align*}
\end{proof}

\subsection{Proof of Theorem~\ref{thm:visits}}
\label{sec:proof-thm-visits}

\begin{proof}
  Since only state $s$ appears below, we will suppress it in the interest of conciseness. Consider
  \begin{align*}
    &\hat{v}_n - v \\
    &\quad\triangleright\text{By \eqref{eq:loop-bellman} and \eqref{eq:loop-estimator} } \\
    &\quad= \left(\hat{\beta}_n + \hat{\alpha}_n\, \hat{v}_n \right) - \left(\beta + \alpha \, v \right) \\
    &\quad\triangleright\text{Rearrange the terms} \\
    &\quad= \hat{\beta}_n - \beta + \hat{\alpha}_n\, \hat{v}_n - \alpha \, v \\
    &\quad\triangleright\text{Add and subtract $\hat{\alpha}_n\, v$} \\
    &\quad= \hat{\beta}_n - \beta + \hat{\alpha}_n\, \hat{v}_n - \hat{\alpha}_n\, v + \hat{\alpha}_n\, v - \alpha \, v \\
    &\quad= \left(\hat{\beta}_n - \beta \right) + \hat{\alpha}_n \left( \hat{v}_n - v \right) + \left( \hat{\alpha}_n - \alpha \right) v. \\
  \end{align*}

  By the definition of an MRP, we have $v \in [0, \nicefrac{r_\text{max}}{1 - \gamma}]$ and $G_1 \in [0, \nicefrac{r_\text{max}}{1 - \gamma})$. Furthermore, $I_1 \geq 1$ implies that $\gamma^{I_1} \in (0, \gamma]$ and $0 < 1 - \gamma \leq 1 - \hat{\alpha}_n$.
  Hence the estimation error is bounded as follows
  \begin{align*}
    \lvert \hat{v}_n - v \rvert &\leq \lvert \hat{\beta}_n - \beta \rvert + \hat{\alpha}_n \, \lvert \hat{v}_n - v \rvert + \lvert \hat{\alpha}_n - \alpha \rvert \, v \\
    \lvert \hat{v}_n - v \rvert - \hat{\alpha}_n \, \lvert \hat{v}_n - v \rvert &\leq \lvert \hat{\beta}_n - \beta \rvert + \lvert \hat{\alpha}_n - \alpha \rvert \, v \\
    &\triangleright\text{Divide by $1 - \hat{\alpha}_n$} \\
    \lvert \hat{v}_n - v \rvert %
    &\leq \left( 1 - \hat{\alpha}_n \right)^{-1} \, \left( \lvert \hat{\beta}_n - \beta \rvert + \lvert \hat{\alpha}_n - \alpha \rvert \, v \right) \\
    &\leq \frac{1}{1 - \gamma} \, \left( \lvert \hat{\beta}_n - \beta \rvert + \lvert \hat{\alpha}_n - \alpha \rvert \, v \right). \\
  \end{align*}

  With failure probability of at most $\delta/2$, from Hoeffding's inequality, we have
  $$ \lvert \hat{\beta}_n - \beta \rvert < \frac{r_\text{max}}{1 - \gamma} \, \sqrt{ \frac{ \log \nicefrac{4}{\delta} }{2 n} } $$
  and similarly
  $$ \lvert \hat{\alpha}_n - \alpha \rvert < \gamma \, \sqrt{ \frac{ \log \nicefrac{4}{\delta} }{2 n} }. $$
  Applying the union bound, we have
  \begin{align*}
    \lvert \hat{v}_n - v \rvert %
    &\leq \frac{1}{1 - \gamma} \, \left( %
      \frac{r_\text{max}}{1 - \gamma} \, \sqrt{ \frac{ \log \nicefrac{4}{\delta} }{2 n} } %
      + %
      \gamma \, v \, \sqrt{ \frac{ \log \nicefrac{4}{\delta} }{2 n} } %
      \right) \\
    &\leq \frac{1}{1 - \gamma} \, \left( %
      \frac{r_\text{max}}{1 - \gamma} \, \sqrt{ \frac{ \log \nicefrac{4}{\delta} }{2 n} } %
      + %
      \gamma \, \frac{r_\text{max}}{1 - \gamma} \, \sqrt{ \frac{ \log \nicefrac{4}{\delta} }{2 n} } %
      \right) \\
    &< \frac{r_\text{max}}{(1 - \gamma)^2} \, \sqrt{ \frac{ \log \nicefrac{4}{\delta} }{2 n} }. &&\qedhere
  \end{align*}
\end{proof}

\subsection{Proof of Lemma~\ref{lem:return-times}}
\label{sec:proof-lem-return-times}
This proof largely follows the proof by \citet{lee2013approximating} and is presented here in the interest of self-containedness.

\begin{proof}
  Suppose $a, b > 0$, consider the probability of the event that $s$ is not visited in the next $a$ steps given that it is not visited in the previous $b$ steps, that is
  \begin{align*}
    &\Prob \left[ H_s^+ > a + b \middle| H_s^+ > b \right] \\
    &\quad\triangleright\text{In particular, $X_b \neq s$} \\
    &\quad \leq \Prob \left[ H_s^+ > a + b \middle| X_b \neq s \right] \\
    &\quad\triangleright\text{By Markov property, we can shift the index} \\
    &\quad = \Prob \left[ H_s^+ > a \middle| X_0 \neq s \right] \\
    &\quad \leq \max_{s' \in \mathcal{S}} \Prob \left[ H_s^+ > a \middle| X_0 = s' \right] \\
    &\quad\triangleright\text{By Markov inequality} \\
    &\quad \leq \max_{s' \in \mathcal{S}} \frac{\Exp \left[ H_s^+ \middle| X_0 = s' \right]}{a} \\
    &\quad\triangleright\text{By definition \eqref{eq:maximal-expected-hitting-time}} \\
    &\quad \leq \frac{\tau_s}{a}. \label{eq:tau-a} \numberthis
  \end{align*}

  Let $a = e\,\tau_s$, and apply the above $\floor*{\frac{t}{a}}$-many times to
  \begin{align*}
    \Prob \sbk*{H_s^+ \geq t} %
    & \leq \Prob \sbk*{ H_s^+ \geq \floor*{\frac{t}{a}}\, a } \\
    &\triangleright\text{Apply \eqref{eq:tau-a}} \\
    & \leq \rbk*{ \frac{\tau_s}{a} }^{\floor*{ \frac{t}{a} }} \\
    &\triangleright\text{Let $a = e\, \tau_s$} \\
    & \leq \rbk*{ \frac{1}{e} }^{\floor*{\frac{t}{e \tau_s}}} \\
    & < e \cdot \rbk*{\frac{1}{e}}^{\frac{t}{e \tau_s}}. &&\qedhere
  \end{align*}
\end{proof}

\subsection{Proof of Corollary~\ref{cor:waiting-times}}
\label{sec:proof-cor-waiting-times}

\begin{proof}
  For conciseness, we suppress $s$ here since only state $s$ appears.
  Suppose $a > 0$. By Remark~\ref{rem:times}, we have $W_n < n\,a$ if $W_1 < a$ and $I_i < a$ for $i = 1, \cdots, n-1$.
  Note that $W_1 \leq H_s^+$ and $I_i$ distribute identically to $H_s^+$. Immediately from inverting Lemma~\ref{lem:return-times}, we have with failure probability of at most $\nicefrac{\delta}{n}$, $W_1$ is bounded
  $$ W_1 \leq H_s^+ < e \tau_s \log \frac{e n}{\delta}. $$
  Suppose each $I_i < a$ fails with probability of at most $\nicefrac{\delta}{n}$, then we similarly have
  $$ I_i < e \tau_s \log \frac{e n}{\delta}. $$
  Applying the union bound, and with probability of at least $1 - \delta$, we have
  $$ W_n < e n \tau_s \log \frac{e n}{\delta}. $$
\end{proof}

\subsection{Proof of Theorem~\ref{thm:steps}}
\label{sec:proof-thm-steps}
\begin{proof}
  First, we introduce the Lambert $\mathcal{W}$ function to invert Corollary~\ref{cor:waiting-times}.
  Recall that the Lambert $\mathcal{W}$ function is a transcendental function defined such that $\mathcal{W}(x) e^{\mathcal{W}(x)} = x$, and thus it is a monotonically increasing function.
  At step $T$, suppose
  \begin{align*}
    e n\, \tau_s \log \frac{e n}{\delta} &= T \\
    \frac{e n}{\delta}\, \log \frac{e n}{\delta} &= \frac{T}{\delta\,\tau_s} \\
    \rbk*{\log \frac{e n}{\delta}}\, e^{\rbk*{\log \frac{e n}{\delta}}} &= \frac{T}{\delta\,\tau_s} \\
    &\triangleright\text{By the definition of $\mathcal{W}$} \\
    \log \frac{e n}{\delta} &= \mathcal{W}\rbk*{\frac{T}{\delta\,\tau_s}} \\
    &\triangleright\text{Exponentiate both sides} \\
    \frac{e n}{\delta} &= e^{\mathcal{W}\rbk*{\frac{T}{\delta\,\tau_s}}}.
  \end{align*}

  Use the fact that if $x > e$, then $\log x - \log\log x < \mathcal{W}(x)$. So given $T > e \delta \tau_s$, we can lower-bound the number of visits
  \begin{align*}
    e^{ \log \rbk*{\frac{T}{\delta\,\tau_s}} - \log\log \rbk*{\frac{T}{\delta\,\tau_s}} } &< \frac{e n}{\delta} \\
    \frac{ \frac{T}{\delta\,\tau_s} }{\log \rbk*{\frac{T}{\delta\,\tau_s}}} &< \frac{e n}{\delta} \\
    \frac{ T\, }{e\, \tau_s\,\log \rbk*{\frac{T}{\delta\,\tau_s}}} &< n.
  \end{align*}

  Plugging this into Theorem~\ref{thm:visits}, we obtain the desired expression
  $$ \vbk*{\hat{v}_T(s) - v(s)} %
  < \frac{r_\text{max}}{(1 - \gamma)^2} \, \sqrt{ \frac{ e \tau_s\,\log \rbk*{\frac{T}{\delta\,\tau_s}} \log \frac{4}{\delta} }{2 T} }. $$
\end{proof}

\subsection{Proof of Corollary~\ref{cor:inf-norm}}
\label{sec:proof-cor-inf-norm}
\begin{proof}
  We run $S$ many copies of loop estimators, one for each state $s \in \mathcal{S}$.
  Following Theorem~\ref{thm:steps}, with failure probability of at most $\nicefrac{\delta}{S}$, we can ensure that each estimator has an error of at most
  $$ \vbk*{\hat{v}_T(s) - v(s)} %
  < \frac{r_\text{max}}{(1 - \gamma)^2} \, \sqrt{ \frac{ e \tau_s\,\log \rbk*{\frac{S\,T}{\delta\,\tau_s}} \log \frac{4 S}{\delta} }{2 T} }. $$

  The largest upper bound comes from the state with the largest maximal expected hitting time $\max_{s \in \mathcal{S}} \tau_s$ of the Markov chain. Apply the union bound and we have
  $$ \dvbk*{\hat{\mathbf{v}}_T - \mathbf{v}}_\infty %
  < \frac{r_\text{max}}{(1 - \gamma)^2} \, \sqrt{ \frac{e \max_s\tau_s\,\log \rbk*{\frac{S\,T}{\delta\,\min_s\tau_s}} \log \frac{4 S}{\delta} }{2 T} }. $$

\end{proof}

\section{Additional results}
\label{sec:additional-results}

\subsection{Conditions for consistency}
\label{sec:final-visit-example}
We provide an example to show that if a state is not positive recurrent, i.e., transient, then we cannot attain a consistent estimate of its value in general. This suggests that Assumption~\ref{assum:reachability} is not too strong as a sufficient condition to study. Recall that we are interested in consistent estimation of the discounted value of a state given a single sample path from an unknown MRP. If a state is not positive recurrent, then without assuming any reset mechanisms, it is visited finitely many times over any sample path almost surely.

\begin{figure}[!t]
  \centering

  \adjustbox{width=0.6\linewidth}{
    \begin{tikzpicture}[->, >=stealth', shorten >=1pt, auto, semithick]
      \tikzstyle{action} = [draw=black,fill=none]
        \node[state, scale=0.9] at (0, 0) (s1) {$s_1'$};
        \node[state, scale=0.9] at (2, 0) (s2) {$s_2$};

        \path (s1) edge [out=0, in=180] node {$1$} (s2);
        \path (s2) edge [out=45, in=-45, looseness=5] node {$1$} (s2);
    \end{tikzpicture}
  }
  %
  %
  \adjustbox{width=0.6\linewidth}{
    \begin{tikzpicture}[->, >=stealth', shorten >=1pt, auto, semithick]
      \tikzstyle{action} = [draw=black,fill=none]
        \node[state, scale=0.9] at (0, 0) (s1) {$s_1$};
        \node[state, scale=0.9] at (2, 1) (s2) {$s_2$};
        \node[state, scale=0.9] at (2, -1) (s3) {$s_3$};

        \path (s1) edge [out=45, in=-135, looseness=0] node {$0.5$} (s2)
                   edge [out=-45, in=135, looseness=0] node {$0.5$} (s3);
        \path (s2) edge [out=45, in=-45, looseness=5] node {$1$} (s2);
        \path (s3) edge [out=45, in=-45, looseness=5] node {$1$} (s3);
    \end{tikzpicture}
  }
  %
  %
  \adjustbox{width=0.6\linewidth}{
    \begin{tikzpicture}[->, >=stealth', shorten >=1pt, auto, semithick]
      \tikzstyle{action} = [draw=black,fill=none]
        \node[state, scale=0.9] at (0, 0) (s1) {$s_1''$};
        \node[state, scale=0.9] at (2, 0) (s3) {$s_3$};

        \path (s1) 
                   edge [out=0, in=180] node {$1$} (s3);
        \path (s3) edge [out=45, in=-45, looseness=5] node {$1$} (s3);
    \end{tikzpicture}
  }
  \caption{Diagram of three Markov reward processes with transition probabilities labeled on the edges. The rewards are $1$ for $s_2$ and $0$ elsewhere.}
  \label{fig:final-visit-example}
\end{figure}
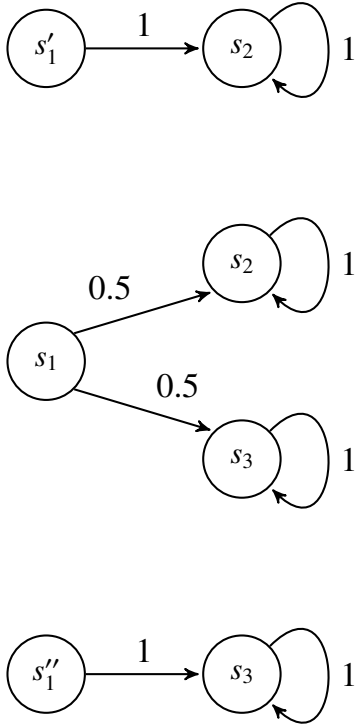

Consider the following three MRPs in Figure~\ref{fig:final-visit-example}. It is obvious that $v(s_1') = \nicefrac{\gamma}{1 - \gamma}$, $v(s_1'') = 0$, and $v(s_1) = \nicefrac{\gamma}{2(1 - \gamma)}$.
Suppose we start in $s_1$ (of the MRP in the middle), there are only two possible sample paths: $(s_1, 0, s_2, 1, s_2, 1, \cdots)$ and $(s_1, 0, s_3, 0, s_3, 0, \cdots)$.
Note that $s_1$ is only visited \emph{once} in either sample path thus transient.
Furthermore, we obtain the first sample path with probability of $\nicefrac{1}{2}$ in which case we cannot distinguish it from a sample path from the MRP on the top. Similarly, with probability of $\nicefrac{1}{2}$, we get the second sample path which is indistinguishable from a sample path from the MRP at the bottom. However, the values of $s_1'$ and $s_1''$ are different (and both not equal to that of $s_1$). Hence we cannot devise an estimator that can \emph{consistently} estimate $v(s_1)$, $v(s_1')$ and $v(s_1'')$.

\subsection{Concentration of first return times}
\label{sec:hitting-time-example}
We provide an example to show that an exponential concentration of first return times given the expected recurrence time is impossible. In contrast, in Lemma~\ref{lem:return-times} we proved an exponential concentration given the maximal expected hitting time. Furthermore, this is consistent with what one would expect from Markov's inequality since first return times are nonnegative random variables.

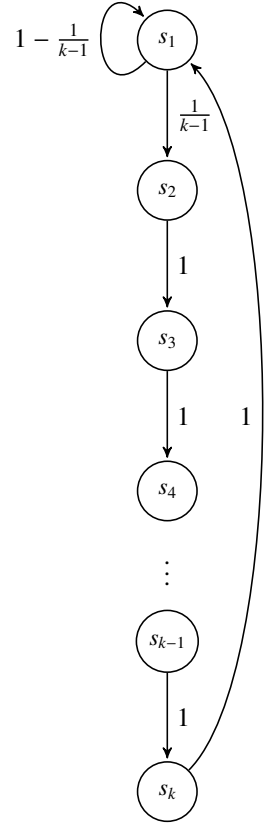
\begin{figure}[!hb]
  \centering
  \begin{tikzpicture}[->, >=stealth', shorten >=1pt, auto, semithick]
    \tikzstyle{action} = [draw=black,fill=none]
      \node[state, scale=0.9] at (0, 0) (s1) {$s_1$};
      \node[state, scale=0.9] at (0, -2) (s2) {$s_2$};
      \node[state, scale=0.9] at (0, -4) (s3) {$s_3$};
      \node[state, scale=0.9] at (0, -6) (s4) {$s_4$};
      \node[state, scale=0.9] at (0, -8) (s5) {$s_{k-1}$};
      \node[state, scale=0.9] at (0, -10) (s6) {$s_k$};

      \path (s1) edge [out=-90, in=90] node {$\frac{1}{k-1}$} (s2)
                 edge [out=-135,in=135,looseness=5] node {$1 - \frac{1}{k-1}$} (s1);
      \path (s2) edge [out=-90, in=90] node {$1$} (s3);
      \path (s3) edge [out=-90, in=90] node {$1$} (s4);
      \path (s4) -- node[auto=false]{$\vdots$} (s5);
      \path (s5) edge [out=-90, in=90] node {$1$} (s6);
      \path (s6) edge [out=45, in=-45,looseness=0.5] node {$1$} (s1);
  \end{tikzpicture}
  \caption{Diagram of Markov chain $M_k$ with transition probabilities labeled on the edges.}
  \label{fig:hitting-time-example}
\end{figure}

Consider a class of Markov chains $\cbk*{M_k}$ indexed by $k \ge 3$ where Markov chain $M_k$ has a state space $\cbk*{s_1, \cdots, s_k}$ and a transition kernel as depicted in Figure~\ref{fig:hitting-time-example}. Starting in $s_1$, the chain $M_k$ can either transition back to $s_1$ in one step with probability $1 - \frac{1}{k - 1}$ or to $s_2$ with probability $\frac{1}{k-1}$.
Thus, there are only two possible values for the first return time to $s_1$: 1 by the self-transition, and $k$ by going through $s_2, s_3, \ldots, s_k, s_1$.
We can calculate the expected recurrence time as
$$ \rho_{s_1} = \rbk*{1 - \frac{1}{k-1}} \cdot 1 + \frac{1}{k-1} \cdot k = 2 . $$
Suppose that there is an exponential concentration of $H^+_{s_1}$ given $\rho_{s_1}$, then we can upper-bound $\Prob\sbk*{H^+_{s_1} \ge t}$ by some exponential function of $t$. However $H^+_{s_1} = k$ with probability of $\frac{1}{k-1}$ in $M_k$ makes such an exponential bound impossible as the upper bound has to work for all $M_k$.

\end{document}